\documentclass[runningheads]{llncs}
\usepackage[T1]{fontenc}
\usepackage{graphicx,verbatim}
\usepackage{multicol}
\usepackage{multirow}

\usepackage{xspace, amsmath, amssymb}
\begin{document}
\title{Graph Laplacian Transformer with Progressive Sampling for Prostate Cancer Grading}

\titlerunning{Graph Laplacian Transformer for Prostate Cancer Grading}

\author{Masum Shah Junayed\inst{1}, John Derek Van Vessem\inst{2}, Qian Wan\inst{2}, Gahie Nam\inst{2}
\and 
Sheida Nabavi\inst{1}
} 
\authorrunning{MS Junayed et al.}

\institute{School of Computing, University of Connecticut (UConn), Storrs, CT 06269, USA \inst{1} \\
Pathology and Laboratory Medicine, UConn Health, Farmington, CT 06030, USA \inst{2}}

\maketitle

\begin{abstract}
Prostate cancer grading from whole-slide images (WSIs) remains a challenging task due to the large-scale nature of WSIs, the presence of heterogeneous tissue structures, and difficulty of selecting diagnostically  relevant regions. Existing approaches often rely on random or static patch selection, leading to the inclusion of redundant or non-informative regions that degrade performance. To address this, we propose a Graph Laplacian Attention-Based Transformer (GLAT) integrated with an Iterative Refinement Module (IRM) to enhance both feature learning and spatial consistency. 
The IRM iteratively refines patch selection by leveraging a pretrained ResNet50 for local feature extraction and a foundation model in no-gradient mode for importance scoring, ensuring only the most relevant tissue regions are preserved. The GLAT models tissue-level connectivity by constructing a graph where patches serve as nodes, ensuring spatial consistency through graph Laplacian constraints and refining feature representations via a learnable filtering mechanism that enhances discriminative histological structures. Additionally, a convex aggregation mechanism dynamically adjusts patch importance to generate a robust WSI-level representation. Extensive experiments on five public and one private dataset demonstrate that our model outperforms state-of-the-art methods, achieving higher performance and spatial consistency while maintaining computational efficiency. 
\end{abstract}
\section{Introduction}
Prostate cancer remains a leading cause of cancer-related mortality worldwide, with whole-slide image (WSI) analysis being essential for grading and risk assessment \cite{bhattacharyya2025efficient}. The Gleason grading system, which evaluates glandular structures, is the standard for prognosis but is challenging due to differences in expert interpretation, high computational costs, and tissue artifacts like folding and staining inconsistencies \cite{linkon2021deep}, \cite{talaat2024improved}. Many computational pathology models treat all tissue patches equally, failing to focus on the most relevant areas \cite{bian2022multiple}. This leads to lower grading accuracy and increased computational burden, highlighting the need for a more efficient and attention based sampling approach \cite{gustafsson2024evaluating}.

Deep learning-based prostate cancer grading predominantly relies on multiple instance learning (MIL) frameworks, where WSI-level labels are inferred from patch-level features. While attention-based MIL models \cite{moranguinho2021attention}, \cite{lin2023interventional}, \cite{lu2021data} attempt to highlight relevant regions, they struggle with non-informative patches due to their reliance on static attention mechanisms, leading to performance degradation. Correlation-based MIL methods \cite{shao2021transmil}, \cite{bian2022multiple}, \cite{myronenko2021accounting} improve inter-patch dependencies but lack explicit spatial constraints, resulting in inconsistent Gleason grading predictions. Graph-based approaches, such as GNN-based models \cite{anklin2021learning}, \cite{behzadi2024weakly}, \cite{pati2023weakly}, build local neighborhood graphs from high-attended patches to capture tissue-level connectivity. However, they require extensive computations and high memory usage, limiting their practicality for real-world applications. Furthermore, transformer-based models \cite{shao2021transmil}, \cite{bontempo2023mil}, \cite{junayed2024scaled}, \cite{jiang2024masked} use self-attention mechanisms to model long-range dependencies, yet they struggle with random patch selection, often discarding critical regions while retaining less informative ones. These challenges highlight the need for a method that adaptively refines patch selection while enforcing spatial constraints to preserve histological consistency.

To overcome these challenges, this work introduces the iterative refinement module (IRM) for adaptive patch selection and graph laplacian attntion-based transfomrer (GLAT) for spatially coherent feature learning. The IRM leverages a pretrained ResNet50 for local feature extraction and a foundation model (FM) \cite{chen2024towards} operating in no-gradient mode to iteratively refine patch importance scores. This ensures that only the most relevant tissue regions contribute while eliminating redundant or non-informative areas. However, IRM does not explicitly model spatial dependencies, which are essential for preserving glandular structures and histological patterns. To address this, the GLAT incorporates graph Laplacian constraints to maintain spatial consistency by modeling histologically similar patches as graph nodes and enforcing smooth feature transitions between them. Additionally, a learnable filtering mechanism refines feature representation by dynamically adjusts the influence of neighboring patches through graph-based feature propagation, 
ensuring spatial coherence while preserving glandular boundaries and tissue morphology. Finally, a convex aggregation mechanism consolidates refined patch features into a robust WSI-level representation, ensuring proportional contribution from the most informative patches for accurate classification. The key contributions of this work are as follows:
\begin{itemize}
\item This work presents a novel transformer-based model that dynamically selects and processes high-relevance regions to improve prostate cancer grading.
\item The Iterative Refinement Module (IRM) introduces an efficient patch selection strategy by refining patch importance scores, eliminating irrelevant regions while reducing computational overhead.
\item The Graph Laplacian Attention-Based Transformer (GLAT) enforces spatial consistency through graph Laplacian constraints and enhances feature representation via learnable filtering.
\item Extensive experiments on five public and one private dataset demonstrate the superiority of the proposed framework over state-of-the-art methods.
\end{itemize}

\section{Proposed Method}

\begin{figure}
\centering
\includegraphics[width=1.0\textwidth]{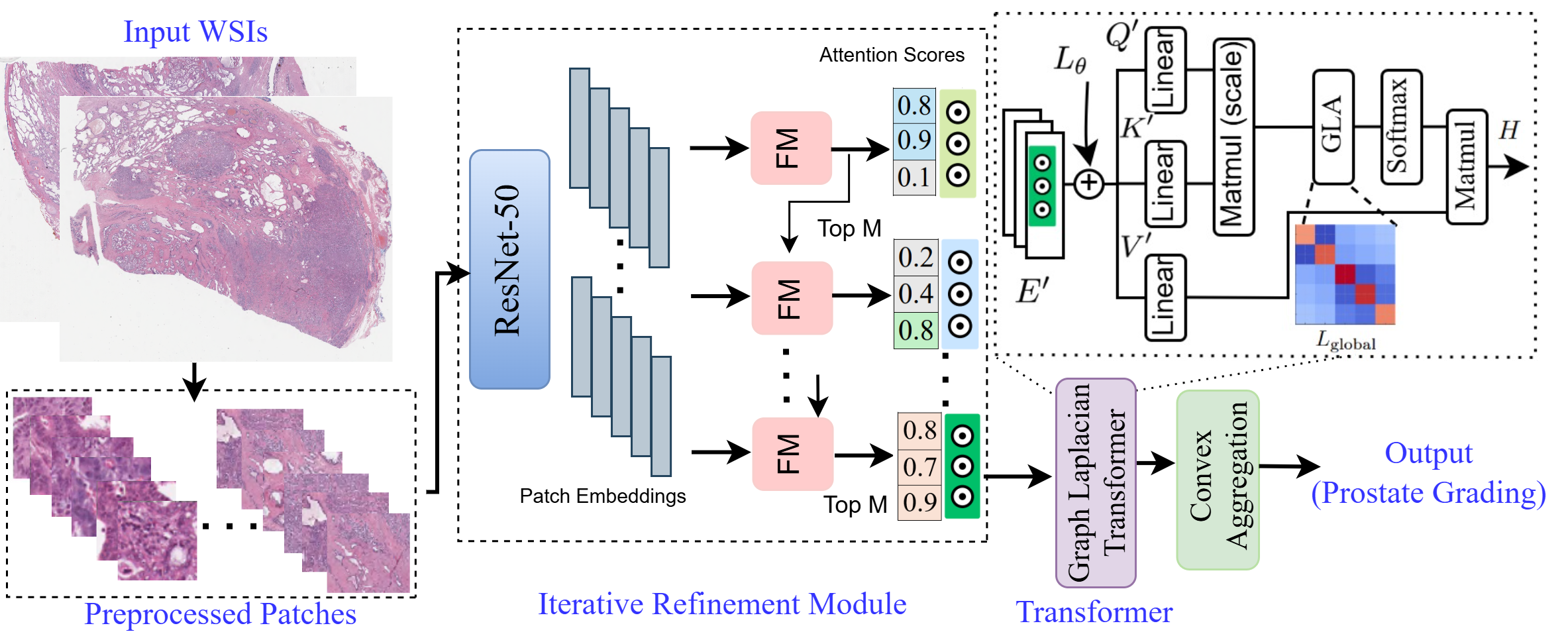}
\caption{Overview of the proposed prostate cancer grading model. The model extracts patches from WSIs, scores their relevance using an IRM with ResNet-50 and a no-gradient FM. These selected patches are then processed through a Graph Laplacian Transformer to capture spatial relationships, followed by convex aggregation for WSI-level feature representation, and a classification head for final Gleason grading.}
\label{fig1}
\end{figure}

Figure~\ref{fig1} depicts the overview of proposed model. As shown in Figure~\ref{fig1}, the model first extracts patch embeddings using a pretrained ResNet50, followed by an IRM to refine high-informative patch selection. Then, Graph Laplacian Transformer to model spatial relationships followed by convex aggregation and classification head for Gleason grading.
\subsection{Iterative Refinement Module}
The IRM operates in two stages: (1) local feature extraction using ResNet50 \cite{he2016deep} and (2) context-aware scoring using a frozen FM, specifically the UNI model~\cite{chen2024towards}. {ResNet50 is used to extract local feature embeddings for each patch, while the FM model, operating in no-gradient mode, assigns attention-based importance scores that capture inter-patch relationships, enabling efficient global reasoning without additional training overhead. The IRM leverages these scores to progressively refine the patch subset across multiple iterations. At each step, patches are rescored based on their contextual relevance, and the least informative ones are discarded. This iterative filtering mechanism allows the model to concentrate on the most diagnostically relevant tissue regions while significantly reducing computational cost.}
Each patch \(P_i \in \mathcal{P}\) is first passed through a pretrained ResNet50 model to extract feature embeddings: \( E_i = f_\text{ResNet}(P_i), \quad E_i \in \mathbb{R}^d,\) where \(f_\text{ResNet}\) represents the ResNet50 feature extractor, and \(d=512\) is the dimensionality of the output embeddings. The embeddings \(E = \{E_1, E_2, \dots, E_N\}\) encapsulate local characteristics of the tissue patches. Next, the embeddings are passed through the FM, which employs a self-attention mechanism to capture pairwise relationships between patches while remaining frozen (i.e., weights are not updated during training). The attention weights \(A_{ij}\) between patches \(i\) and \(j\) are computed as:
\begin{equation}
A_{ij} = \text{softmax}\left(\frac{Q_i K_j^\top}{\sqrt{d_k}}\right), \quad Q_i = W_Q E_i, \, K_j = W_K E_j,
\end{equation}
where \(W_Q, W_K \in \mathbb{R}^{d \times d_k}\) are fixed projection matrices from the FM that define the query and key representations, ensuring that the learned attention mechanism is based on pretrained knowledge. The softmax function normalizes the attention scores, ensuring that the influence of each patch is effectively distributed across all other patches.

Using the computed attention weights, the FM refines the embeddings ($E_i'$) of each patch by aggregating information from its neighbors: \( E_i' = \sum_{j=1}^N A_{ij} V_j,\) \(V_j = W_V E_j,\) where \(W_V \in \mathbb{R}^{d \times d_v}\) projects the value vectors. The refined embeddings \(E'=\{E_1', E_2', \dots, E_N'\}\) encode both local features and global contextual dependencies, making them suitable for scoring the patches. These embeddings are then used to compute patch importance scores: \(S_i = \frac{\sum_{j=1}^N A_{ij}}{N},\) where \(S_i\) represents the average attention weight of patch \(P_i\) across all other patches, reflecting its overall contribution to the contextual structure of the WSI.

Patch embeddings are divided in several ($T$) non-overlapping subsets. At the first iteration (\(t = 0\)), the first subset of  patch embeddings are passed through the FM, which assigns initial importance scores \(S_i^{(0)}\). Based on these scores, the top \(M\) patches with the highest scores are selected: \(\mathcal{P}_\text{selected}^{(0)} = \{P_i^{(0)} : S_i^{(0)} \text{ is among the top } M\}.\)
The subset of selected patches are denoted as \(\mathcal{P}_\text{selected}^{(0)} = \{P_{i_1}^{(0)}, P_{i_2}^{(0)}, \dots, P_{i_{M}}^{(0)}\}\). In each subsequent iteration \(t\), the embeddings of the selected patches from the previous iteration are combined with the next subset of patches and reprocessed using the FM, which recalculates their contextual relationships with the other patches in the subset. The refined embeddings\(\{E_i'^{(t)}\}\), and scores \(\{S_i^{(t)}\}\) are updated as :
\begin{equation}
E_i'^{(t)} = \sum_{j \in {N/T}} A_{ij}^{(t)} V_j^{(t)}, \quad S_i^{(t)} = \frac{\sum_{j=1}^{N/T} A_{ij}^{(t)}}{{N/T}}.
\end{equation}
At each iteration, the FM updates the patch importance scores, progressively refining the selection process by keeping only the top 
\(M\) patches with the highest scores: \(\mathcal{P}_\text{selected}^{(t)} = \{P_i^{(t)} : S_i^{(t)} \text{ is among the top } M\}.\)
At the end of \(T\) iterations, the IRM process produces the final set of selected patches \(\mathcal{P}_\text{selected}^{(T)}\) is passed to the next stage of the framework for downstream analysis.

\subsection{Graph Laplacian Transformer}
{To capture both spatial coherence and long-range contextual dependencies among high-informative patches, the GLAT is introduced. The GLAT addresses a critical limitation of standard multihead self attention (MSA) by explicitly enforcing spatial consistency. Unlike MSA, which lacks spatial regularization, GLAT models spatial and morphological relationships by connecting histologically similar patches in a graph structure.}
To explicitly model the spatial relationships and tissue-level connectivity among selected patches, we represent them as a node in a graph \( G = (Vt, X) \). Here, \( Vt \) corresponds to the high-informative patches and \( X \) defines the edges that capture histological feature similarity.
To construct the graph, an edge \(X_{ij}\) is established between patches \(i\) and \(j\) based on their feature similarity. The adjacency matrix \( W \) is computed using a Gaussian kernel function to quantify the pairwise similarity between patches:
\begin{equation}
W_{ij} = \exp\left(-\frac{\|E_i' - E_j'\|^2}{2\sigma^2}\right),
\end{equation}
where \( W_{ij} \) measures the feature similarity between patches \( i \) and \( j \), with \( E_i' \) and \( E_j' \) denoting their respective feature embeddings. The parameter \( \sigma \) controls the sensitivity of similarity weighting. The degree matrix \( D \), defined as : \( D_{ii} = \sum_{j} W_{ij} \), Using these matrices, the global graph Laplacian is computed as: \( L_{\text{global}} = D - W \).

To refine feature representations before self-attention, GLAT employs a learnable filtering mechanism that dynamically adjusts feature propagation across patches. This is formulated as: \( Q' = L_{\theta} Q, \quad K' = L_{\theta} K, \quad V' = L_{\theta} V,\) where \( L_{\theta} \) is a trainable filter optimized during training to control feature propagation. 
\( L_{\theta} \) learns adaptive transformations that enhance discriminative features while preserving local structural details. Using the learnable-filtered queries, keys, and values, the modified graph laplacian attention (GLA) mechanism is computed as:
\begin{equation}
A' = \text{softmax} \left(\frac{Q' K'^\top + \lambda L_{\text{global}}}{\sqrt{d_k}}\right),
\end{equation}
where \( \lambda \) is a tunable hyperparameter that regulates the influence of the spatial constraints, ensuring an optimal balance between feature-driven attention and structured spatial coherence within the GLA mechanism.. The resulting attention scores refine feature embeddings as: \( H = A' V', \quad H \in \mathbb{R}^{M \times d},\) where \( H \) contains individual refined embeddings for each selected patch. 
To generate a global WSI representation, we apply convex aggregation \cite{iso2021convex}, which ensures that the most relevant refined patches contribute proportionally:
\begin{equation}
H_\text{WSI} = \sum_{i=1}^M w_i H_i', \quad w_i = \frac{\exp(\theta_i)}{\sum_{j=1}^{M} \exp(\theta_j)},
\end{equation}
where \(\theta_i\) are trainable parameters that determine the relative importance of each patch. The softmax function is applied to \(\theta_i\) to obtain normalized weights \(w_i\), ensuring that they are non-negative \( w_i \geq 0, \text{and} \sum_{i=1}^M w_i = 1. \) This normalization allow the model to dynamically adjust patch importance during training while maintaining a balanced feature aggregation. Finally, the WSI representation \( H_{\text{WSI}} \) is passed through a classification head: \(y = \text{Softmax}(\text{Linear}(H_{\text{WSI}})).\)

The model is trained using categorical cross-entropy loss for Gleason grading. To further encourage spatial consistency, a Graph-based feature smoothness constraint is incorporated into the loss function:
\begin{equation}
\mathcal{L}_{\text{total}} = \mathcal{L}_{\text{CE}} + \alpha \sum_{i,j} W_{ij} \|H_i - H_j\|^2,
\end{equation}
where \( \mathcal{L}_{\text{CE}} \) represents the loss of standard classification, and the second term encourages the consistency of characteristics between spatially similar patches. The hyperparameter \( \alpha \) balances classification performance with spatial coherence.
\section{Experimental Results}

\subsection{Datasets and Preprocessing}
This study utilizes five diverse publicly availabe datasets for evaluating the proposed framework: TCGA-PRAD \cite{goldman2020visualizing}, SICAPv2 \cite{silva2020going}, GLEASON19 \cite{nir2018automatic}, PANDA, DiagSet \cite{koziarski2024diagset}, and a Private dataset. The private (UConn Health) dataset includes 79 WSIs, enhancing the evaluation of the model on a smaller yet clinically relevant dataset. For grading, the ISUP classification system was employed, categorizing samples into four classes: Grade 1 and 2 representing normal tissue, and Grades 3, 4, and 5 indicating varying levels of malignancy. For preprocessing, the CLAM \cite{lu2021data} method was employed to generate high-quality patches from WSIs and TMAs. The preprocessing pipeline included stain normalization, tissue segmentation, and patch extraction with a fixed size, ensuring consistency across datasets. Patches with minimal tissue content were excluded to enhance data quality, and each patch was normalized to reduce staining variability.

\subsection{Experimental Setup}
All models, including the proposed method and existing baseline methods, were trained on high-performance GPUs (NVIDIA RTX A6000) to handle large-scale histopathological datasets. 
The input patches were extracted using the CLAM \cite{lu2021data} standard preprocessing pipeline with a patch size of 224, followed by standard data augmentation techniques such as random flipping and rotation to enhance model robustness. A batch size of 16 was used with an initial learning rate of \(1 \times 10^{-4}\), optimized using the Adam optimizer with a weight decay set to \(1 \times 10^{-5}\). The early stopping strategy was applied to prevent overfitting based on the validation performance, and all models were trained for up to 100 epochs. The performance of the model was assessed primarily using AUC and Cohen’s Kappa (CK) reported as mean over five-fold cross-validation for statistical reliability. 
The proposed method was evaluated against state-of-the-art baseline models. To ensure fair comparison, publicly available codebases were used, and all hyperparameters were aligned with the original implementations.

\subsection{Results and Discussions}
\begin{table}[]
\centering
\caption{Quantitative comparison of the proposed method against state-of-the-art approaches on six prostate cancer grading datasets. Performance is measured using AUC and CK. The best results are in bold, and the second-best results are underlined.}
\scalebox{0.88}{
\begin{tabular}{|c|cc|cc|cc|cc|cc|cc|}
\hline
\multirow{2}{*}{\begin{tabular}[c]{@{}c@{}}Dataset/\\ Methods\end{tabular}} & \multicolumn{2}{c|}{SICAPv2} & \multicolumn{2}{c|}{TCGA-PRAD} & \multicolumn{2}{c|}{GLESON19} & \multicolumn{2}{c|}{PANDA} & \multicolumn{2}{c|}{Diagset} & \multicolumn{2}{c|}{Private} \\ \cline{2-13} 
 & \multicolumn{1}{c|}{AUC} & CK & \multicolumn{1}{c|}{AUC} & CK & \multicolumn{1}{c|}{AUC} & CK & \multicolumn{1}{c|}{AUC} & CK & \multicolumn{1}{c|}{AUC} & CK & \multicolumn{1}{c|}{AUC} & CK \\ \hline
ABMIL \cite{moranguinho2021attention} & \multicolumn{1}{c|}{0.658} & 0.598 & \multicolumn{1}{c|}{0.616} & 0.591 & \multicolumn{1}{c|}{0.648} & 0.601 & \multicolumn{1}{c|}{0.631} & 0.605 & \multicolumn{1}{c|}{0.598} & 0.546 & \multicolumn{1}{c|}{0.534} & 0.496 \\ \hline
TranMIL \cite{shao2021transmil} & \multicolumn{1}{c|}{0.593} & 0.567 & \multicolumn{1}{c|}{0.587} & 0.538 & \multicolumn{1}{c|}{0.612} & 0.582 & \multicolumn{1}{c|}{0.605} & 0.580 & \multicolumn{1}{c|}{0.535} & 0.486 & \multicolumn{1}{c|}{0.481} & 0.437 \\ \hline
MST \cite{bian2022multiple} & \multicolumn{1}{c|}{0.918} & 0.796 & \multicolumn{1}{c|}{0.863} & 0.792 & \multicolumn{1}{c|}{0.839} & 0.810 & \multicolumn{1}{c|}{0.895} & 0.851 & \multicolumn{1}{c|}{0.720} & 0.679 & \multicolumn{1}{c|}{0.679} & 0.609 \\ \hline
DASMIL \cite{bontempo2023mil} & \multicolumn{1}{c|}{0.915} & 0.819 & \multicolumn{1}{c|}{0.867} & 0.799 & \multicolumn{1}{c|}{0.846} & 0.808 & \multicolumn{1}{c|}{0.897} & 0.846 & \multicolumn{1}{c|}{0.736} & 0.685 & \multicolumn{1}{c|}{0.683} & 0.628 \\ \hline
WSDMPC \cite{behzadi2024weakly} & \multicolumn{1}{c|}{0.819} & 0.785 & \multicolumn{1}{c|}{0.835} & 0.805 & \multicolumn{1}{c|}{0.826} & 0.808 & \multicolumn{1}{c|}{0.860} & 0.806 & \multicolumn{1}{c|}{0.678} & 0.618 & \multicolumn{1}{c|}{0.650} & 0.605 \\ \hline
MaskHIT \cite{jiang2024masked} & \multicolumn{1}{c|}{0.938} & 0.868 & \multicolumn{1}{c|}{0.909} & 0.856 & \multicolumn{1}{c|}{0.887} & 0.853 & \multicolumn{1}{c|}{0.922} & 0.901 & \multicolumn{1}{c|}{0.785} & 0.724 & \multicolumn{1}{c|}{0.746} & 0.701 \\ \hline
SMAHM \cite{junayed2024scaled} & \multicolumn{1}{c|}{0.941} & 0.881 & \multicolumn{1}{c|}{0.918} & 0.876 & \multicolumn{1}{c|}{\textbf{0.914}} & 0.876 & \multicolumn{1}{c|}{\underline{ 0.958}} & \underline{ 0.913} & \multicolumn{1}{c|}{\underline{ 0.819}} & \underline{ 0.768} & \multicolumn{1}{c|}{\underline{ 0.755}} & \underline{ 0.719} \\ \hline
HEAT \cite{chan2023histopathology} & \multicolumn{1}{c|}{\underline{ 0.943}} & \underline{ 0.875} & \multicolumn{1}{c|}{\underline{ 0.921}} & \textbf{0.886} & \multicolumn{1}{c|}{0.905} & \underline{ 0.889} & \multicolumn{1}{c|}{0.946} & 0.909 & \multicolumn{1}{c|}{0.805} & 0.759 & \multicolumn{1}{c|}{0.748} & 0.706 \\ \hline
Proposed & \multicolumn{1}{c|}{\textbf{0.951}} & \textbf{0.892} & \multicolumn{1}{c|}{\textbf{0.923}} & \underline{0.885} & \multicolumn{1}{c|}{\underline{ 0.913}} & \textbf{0.892} & \multicolumn{1}{c|}{\textbf{0.963}} & \textbf{0.936} & \multicolumn{1}{c|}{\textbf{0.836}} & \textbf{0.791} & \multicolumn{1}{c|}{\textbf{0.781}} & \textbf{0.731} \\ \hline
\end{tabular}}
\label{tab:qn}
\end{table}
Table \ref{tab:qn} presents the performance comparison of the proposed method with existing state-of-the-art approaches on six benchmark datasets, including SICAPv2, TCGA-PRAD, GLEASON19, PANDA, DiagSet, and Private. The evaluation metrics used are AUC and Cohen’s Kappa, both of which assess model reliability and agreement with ground truth annotations. The proposed method consistently achieves the highest AUC and Kappa scores across all datasets, demonstrating superior grading performance. Notably, the model outperforms the closest competitor, HEAT, on SICAPv2, TCGA-PRAD, and PANDA, while maintaining competitive performance on the remaining datasets. The significant improvement in CK highlights the model’s robustness in handling class imbalance and variability in histopathological features.

\begin{table}[]
\centering
\caption{Ablation study evaluating the impact of key components in the proposed framework for prostate cancer grading.}
\begin{tabular}{|c|ccc|ccc|c|cc|cc|}
\hline
\multirow{2}{*}{Exp.} & \multicolumn{3}{c|}{IRM} & \multicolumn{3}{c|}{Transformer} & \multirow{2}{*}{CA} & \multicolumn{2}{c|}{Performance} & \multicolumn{2}{c|}{Comp. Cost} \\ \cline{2-7} \cline{9-12} 
 & \multicolumn{1}{c|}{ResNet50} & \multicolumn{1}{c|}{FM} & IP & \multicolumn{1}{c|}{GLA} & \multicolumn{1}{c|}{MSA} & SWA &  & \multicolumn{1}{c|}{AUC} & CK & \multicolumn{1}{c|}{\#P(M)} & Flops \\ \hline
1 & \multicolumn{1}{c|}{\textbf{\checkmark}} & \multicolumn{1}{c|}{\textbf{\checkmark}} & \textbf{\checkmark} & \multicolumn{1}{c|}{\textbf{\checkmark}} & \multicolumn{1}{c|}{\textbf{$\times$}} & \textbf{$\times$} & \checkmark & \multicolumn{1}{c|}{\textbf{0.781}} & \textbf{0.731} & \multicolumn{1}{c|}{\textbf{83.3}} & \textbf{32.53} \\ \hline
2 & \multicolumn{1}{c|}{$\times$} & \multicolumn{1}{c|}{\checkmark} & \checkmark & \multicolumn{1}{c|}{\checkmark} & \multicolumn{1}{c|}{$\times$} & $\times$ & \checkmark & \multicolumn{1}{c|}{0.779} & 0.725 & \multicolumn{1}{c|}{89.9} & 43.56 \\ \hline
3 & \multicolumn{1}{c|}{\checkmark} & \multicolumn{1}{c|}{$\times$} & \checkmark & \multicolumn{1}{c|}{\checkmark} & \multicolumn{1}{c|}{$\times$} & $\times$ & \checkmark & \multicolumn{1}{c|}{0.737} & 0.696 & \multicolumn{1}{c|}{98.6} & 49.46 \\ \hline
4 & \multicolumn{1}{c|}{\checkmark} & \multicolumn{1}{c|}{\checkmark} & $\times$ & \multicolumn{1}{c|}{\checkmark} & \multicolumn{1}{c|}{$\times$} & $\times$ & \checkmark & \multicolumn{1}{c|}{0.763} & 0.708 & \multicolumn{1}{c|}{130.5} & 91.63 \\ \hline
5 & \multicolumn{1}{c|}{\checkmark} & \multicolumn{1}{c|}{\checkmark} & \checkmark & \multicolumn{1}{c|}{$\times$} & \multicolumn{1}{c|}{\checkmark} & $\times$ & \checkmark & \multicolumn{1}{c|}{0.754} & 0.709 & \multicolumn{1}{c|}{86.7} & 41.58 \\ \hline
6 & \multicolumn{1}{c|}{\checkmark} & \multicolumn{1}{c|}{\checkmark} & \checkmark & \multicolumn{1}{c|}{$\times$} & \multicolumn{1}{c|}{$\times$} & \checkmark & \checkmark & \multicolumn{1}{c|}{0.751} & 0.710 & \multicolumn{1}{c|}{84.2} & 33.9 \\ \hline
7 & \multicolumn{1}{c|}{\checkmark} & \multicolumn{1}{c|}{\checkmark} & \checkmark & \multicolumn{1}{c|}{\checkmark} & \multicolumn{1}{c|}{$\times$} & $\times$ & $\times$ & \multicolumn{1}{c|}{0.769} & 0.662 & \multicolumn{1}{c|}{84.1} & 35.94 \\ \hline
\end{tabular}
\label{tab:abl}
\end{table}
Table \ref{tab:abl} presents an ablation study assessing the impact of individual components within the proposed framework on Private dataset. The full model (Exp. 1), incorporating IRM with GLA and CA, achieved the highest performance (AUC: 0.781, CK: 0.731) with an optimal computational balance (83.3M parameters, 32.53 FLOPs). Replacing ResNet50 with ViT \cite{dosovitskiy2020image} (Exp. 2) slightly reduced performance while increasing computational cost. Excluding FM (Exp. 3) significantly lowered AUC (0.737) and CK (0.696), demonstrating its importance in refining patch selection. Removing iterative process (IP) (Exp. 4) and relying on random patch selection led to degraded performance and increased overhead (130.5M parameters, 91.6 FLOPs), emphasizing the necessity of adaptive sampling. Substituting GLA with MSA (Exp. 5) or SWA (Exp. 6) resulted in lower AUC, confirming the advantage of GLA in modeling spatial relationships. Lastly, replacing CA with mean pooling (Exp. 7) significantly reduced AUC (0.969 to 0.662), highlighting CA’s role in forming a robust WSI representation.
\begin{figure}
\centering
\includegraphics[width=0.9\textwidth]{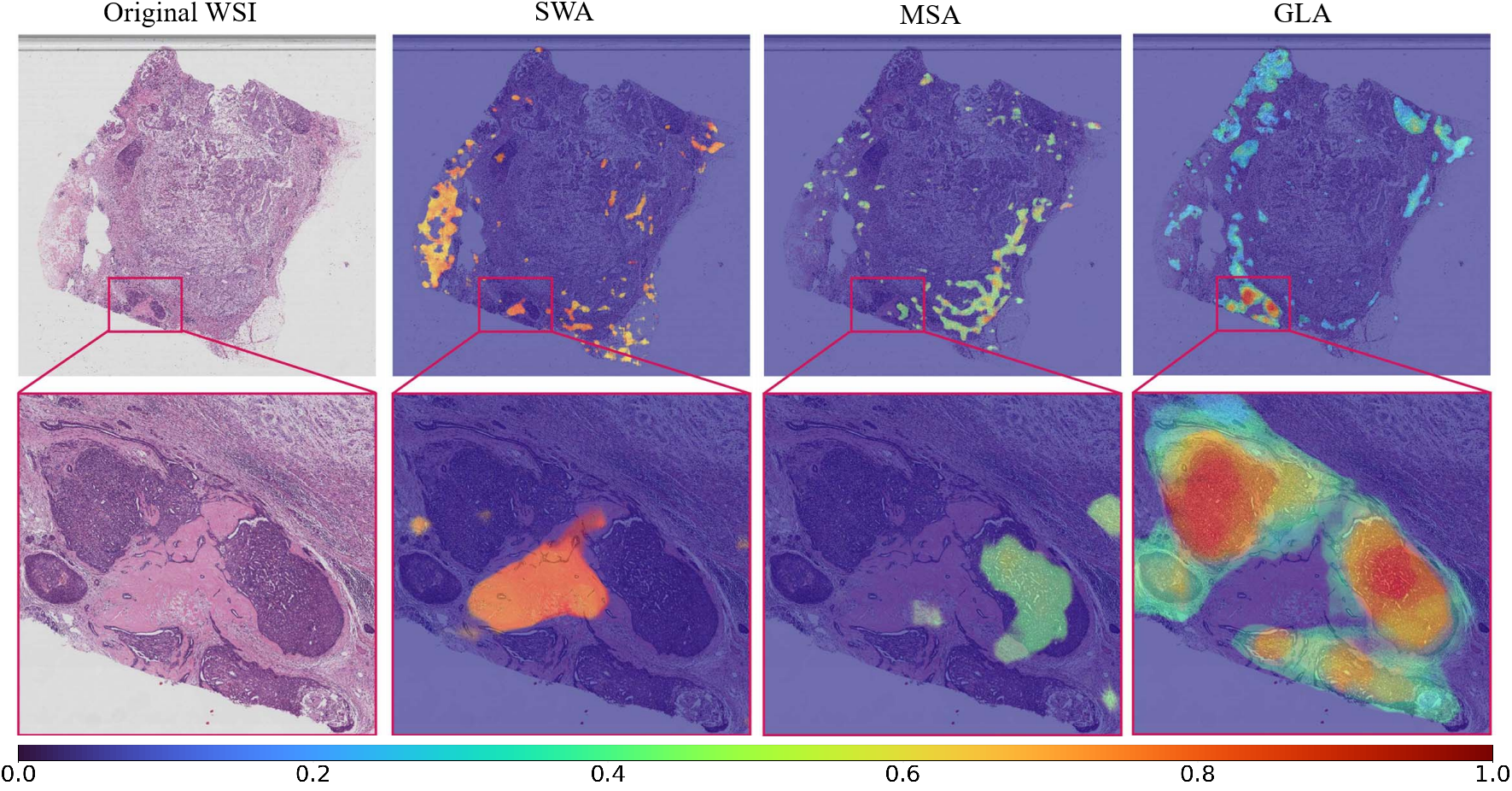}
\caption{Visualization of attention score maps for different mechanisms in prostate cancer grading.
} \label{fig2}
\end{figure}
Figure \ref{fig2} presents a comparative analysis of different attention mechanisms— shifted window attention (SWA) \cite{liu2021swin}, MSA \cite{dosovitskiy2020image}, and GLA—for prostate cancer grading. The first column displays the original WSI. The second, third, and fourth columns illustrate the attention heatmaps generated by SWA, MSA, and GLA, respectively. The magnified views in the second row further highlight the differences in feature localization. SWA captures localized features but lacks a global perspective, resulting in fragmented attention. MSA expands attention to broader regions but does not enforce spatial consistency, sometimes misaligning critical areas. In contrast, GLA integrates graph Laplacian constraints to maintain spatial coherence across histologically similar regions while refining feature embeddings. Expert pathologists confirmed that the highlighted areas in MSA and GLA correspond to cancerous regions, validating the effectiveness of our approach. Unlike SWA and MSA, GLA preserves high-frequency histological structures , such as glandular boundaries and morphological variations, by applying learnable filtering at the feature level. The heatmaps demonstrate that GLA provides a more structured and precise attention distribution, highlighting the importance of spatial constraints for accurate prostate cancer grading.

\section{Conclusions and Future Work}
This work presents a novel framework combining an Iterative Refinement Module (IRM) for adaptive patch selection and a Graph Laplacian Attention-Based Transformer (GLAT) for spatially coherent feature learning, achieving state-of-the-art performance in prostate cancer grading. By refining patch selection and enforcing spatial constraints, the model effectively captures histopathological structures while reducing computational overhead. Extensive experiments on public and private datasets validate its effectiveness and generalizability. However, the framework relies on fixed patch sizes, which may not fully capture multi-scale tissue variations. Future work will explore adaptive patch selection and cross-slide attention mechanisms to enhance contextual awareness across distant tumor regions.
\subsubsection{Acknowledgment :}This work is supported in part by the National Science Foundation (NSF) under grant No. 2348278, PI: Nabavi.

\bibliographystyle{splncs04}

\bibliography{strings, Paper5209}
\end{document}